\documentclass[conference]{IEEEtran}
\IEEEoverridecommandlockouts
\usepackage{cite}
\usepackage{amsmath,amssymb,amsfonts}
\usepackage{algorithmic}
\usepackage{graphicx}
\usepackage{textcomp}
\usepackage{xcolor}
\usepackage{tabularx}
\usepackage{booktabs}
\usepackage{mathrsfs}
\usepackage{mathtools}
\usepackage{subfig}
\def\BibTeX{{\rm B\kern-.05em{\sc i\kern-.025em b}\kern-.08em
    T\kern-.1667em\lower.7ex\hbox{E}\kern-.125emX}}
\begin{document}
\newcommand\Tstrut{\rule{0pt}{2.0ex}}       
\newcommand\Bstrut{\rule[-0.9ex]{0pt}{0pt}} 
\title{ Task-Agnostic Federated Learning with Imbalanced Data
}

\author{
  \IEEEauthorblockN{
    Zhengtao Yao\IEEEauthorrefmark{1}\IEEEauthorrefmark{2},\quad
    Hong Nguyen\IEEEauthorrefmark{1}\IEEEauthorrefmark{2},\quad
    \thanks{\IEEEauthorrefmark{1} Corresponding authors}
    Ajitesh Srivastava\IEEEauthorrefmark{2},\quad
    Jose Luis Ambite\IEEEauthorrefmark{2},\quad
  }
  \IEEEauthorblockA{
    \IEEEauthorrefmark{2} University of Southern California, Los Angeles, CA 90089\\
    {hongn}@usc.edu
  }
}

\maketitle

\begin{abstract}
In the realm of medical imaging, leveraging large-scale datasets from various institutions is crucial for developing precise deep learning models, yet privacy concerns frequently impede data sharing. Federated learning (FL) emerges as a prominent solution for preserving privacy while facilitating collaborative learning. 
However, its application in real-world scenarios faces several obstacles, such as task \& data heterogeneity, label scarcity, non-identically distributed (non-IID) data, computational variation, etc.  In real-world, medical institutions may not want to disclose their tasks to FL server and generalization challenge of out-of-network institutions with un-seen task want to join the on-going federated system.  This study addresses task-agnostic and generalization problem on unseen tasks by adapting self-supervised FL framework. Utilizing Vision Transformer (ViT) as consensus feature encoder for self-supervised pre-training, no initial labels required, the framework enables effective representation learning across diverse datasets and tasks. Our extensive evaluations, using various real-world non-IID medical imaging datasets, validate the approach's efficiency. The proposed model retains 90\% of F1 accuracy of centralized approaches for classification task, and outperforms it for segmentation task, which exhibits adaptability to out-of-distribution data. The result indicates that federated learning architecture can be a potential approach toward multi-task foundation modeling.
\end{abstract}


\section{Introduction}
Federated learning (FL) enables model training using data dispersed across multiple locations without direct data sharing. In comparison to models trained at individual sites, federated models can utilize a significantly broader and larger dataset, potentially leading to enhanced performance and greater generalizability. As a result, this training method has found widespread adoption in crucial medical applications such as brain tumor detection and COVID-19 diagnosis, and has been utilized across various data types. 
However, 
In conventional FL system, the center model has to know what task each sites want to achieve, e.g., predict brain ages, detect tumor, classify Diabetic retinopathy, brain regional segmentation, etc. Thus, the task-agnosticity limits the generalization of FL system.
\begin{figure}[t]
  \centering
    \includegraphics[width=\linewidth]{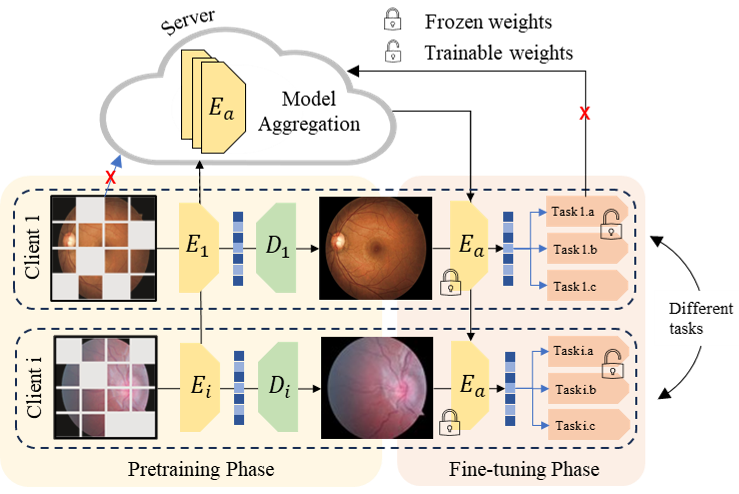}
    \caption{Federated Learning system where clients' tasks are anonymous between each others and hidden from the server. $E_i$ is the $i_{th}$ feature encoder and $D_i$ is the $i_{th}$ decoder.}
    \label{fig:system2}
\end{figure}


Recently, advantage in unsupervised and self-supervised learning narrows the gap in understanding medical images' context, modality without knowing the tasks. Self-supervised pre-training learns the intrinsic features of images in local clients without labels, embodying less label-specific inductive bias and thus, less susceptible to label distribution skewness. The proposed method learns visual representations more effectively across non-IID clients, even when data are limited at some clients. No task was reveal to server and still, pretrained model learn semantic information of domain knowledge which can be used for few-shot learning at each FL sites.

To the best of authors' knowledge, many related works assume that the center model knows all, or at least some tasks and possible labels of local clients\cite{b32}\cite{b33}. We address the challenges of data heterogeneity and tasks anonymous by proposing a novel SSL-FL framework, following the unsupervised federated learning setting in \cite{b15}. It tackles the non-iid data issue using vision mamba to  employ masked image modeling as the self-supervised task, which are efficient in processing long sequences non-iid data because of its convolutional and near-linear computation. The proposed pre-training scheme significantly advances the capability of federated models over
highly heterogeneous data partitions. We also propose a novel client-specific fine tuning framework using low rank adaptation, enabling the existing client to achieve high prediction performance by fintuning only 2\% of the pretrained model. 
Moreover, We show that our framework is robust to merging new institutes as well as new task. 
Our main contribution are summarized as follows:

\begin{itemize}
    \item We define a new problem setting in federated learning involving task-agnostic between clients and server. Along with that, we proposed a simple yet powerful approach to address task-agnostic challenges at FL clients using self-supervised pretraining. After the pretraining, we are able to transfer the knowledge to the target tasks by a efficient fine-tuning implementation.
    \item Using real-world datasets, we show that the combined framework robust to multiple tasks as it achieves up 90\% of SSL central training performance in term of F1 and accuracy with only their own data and outperforms the central model in segmentation tasks. It is found in the experiments that tasks with less data benefits more from federated pretraining.
    
\end{itemize}
\section{Related Works}
\noindent
\textbf{Client-Heterogeneity in Federated Learning.} As a decentralized approach, FL suffers from performance degradation due to client heterogeneity \cite{b37}. 
While several research efforts \cite{b15} have
been devoted to addressing the challenges caused by data
heterogeneity, task heterogeneity have not been well investigated in literature. 
Moreover, the success of such models largely relies on supervised ImageNet pre-training, which could suffer from domain discrepancy when fine-tuning with medical images and can be further improved by self-supervised pre-training on a centrally shared large-scale in-domain medical dataset \cite{b15}. 
However, such centrally shared datasets rarely exist in the medical domain due to privacy and ownership concerns. Therefore, it is desired to build a self-supervised FL framework that collaboratively learns a global model by leveraging all available unlabeled data without sharing data among institutions.

\noindent
\textbf{Vision Foundation model and fine-tuning} Understand medical images without labels is a long-standing problem. At current state, label deficiency is a common challenge in medical imaging.
To address this issue, various approaches such as semi-supervised and self-supervised learning methods \cite{b38}\cite{b39}\cite{b40}\cite{b41} have been proposed to allow models to learn from partially labeled or unlabeled data. This work make use of advantage in SSL on FL systems for task-agnostic settings. 

Specifically, large model fine-tuning and the use of adapters are evolving learning methods and pivotal strategies in the field of large language model\cite{b16} \cite{b17}, vision model\cite{b18} \cite{b19}, and vision language model\cite{b20} \cite{b21}. These approaches are designed to tailor pre-trained models to specific tasks or datasets with relatively minimal computational cost and data requirements.  

\noindent

\section{Problem Statement}


Suppose there are $\mathcal{N}$ distributed clients and one centralized server in a federated system. Each client has a private dataset $\mathcal{D}$ and performs multiple private tasks $\mathcal{T}=\{\mathcal{T}_1,... \mathcal{T}_k\}$ with the dataset they have. The server neither have access to $\mathcal{D}$ nor knows the underlining tasks. Take advantage of un-labeled data from multi-modality and anonymous tasks at clients, we want to boost the performance of each clients compare with local supervision. 
There are similar works on multi-task federated learning that aggregate observed tasks but the server must know beforehand every possible tasks of clients.
With identical SSL-FL architecture, address non-IID and data heterogeneity problem, on a single task.
Unlike any previous works, we introduce a strictly constrained problem involved task-related leakage concerns at federated server.

To address this problem, we assume that there is a unified block that is compromised among clients and server, which have to be feature encoder. In this work, we pre-trained the feature encoder at clients and aggregate it on server to generalize the feature space with out-of-distribution data and tasks.
\section{Methodology}
\subsection{SSL-FL Pre-training}
During the pre-training phase, each local model, indexed by $i$, functions as an autoencoder with an encoder $E_i$ and a decoder $D_i$. This involves masked image modeling, where some image patches are masked and the original signals in these patches are reconstructed \cite{b34}. The encoder in a Vision Transformer processes image patches using multiple layers of self-attention and feed-forward networks to encode complex features and contextual relationships. The decoder reconstructs the original image from these encoded features, utilizing similar layers to generate predictions.
Each local encoder $E_i$ and decoder $D_i$ are trained with local data $D^i$ to minimize the local objective function $L_i(w)=\mathbb{E}_{x\sim\mathcal{X}^i}\big[l_k\big(w;\boldsymbol{x}\big)\big]$, where $l_i$ is the mean squared error of the predicted pixel values of the masked patches. The local loss function $l_k$ is given by:
 
$$l_k=\sum_{j\in\mathcal{P}}\frac1{|\mathcal{P}|}((\boldsymbol{x}_p^j-\hat{\boldsymbol{x}}_p^j)^2;w)$$

For pre-training, each local client update its local model $E_k$ and $D_k$ by minimizing its own loss $L_k$ on data $\mathcal{X}^k.$ Then, the server takes a weighted average of all the resulting local models to update the global model $E_a$ and $D_a$,which is further sent back to the local clients for the next training iteration. 
Once pre-training is complete, the final pre-trained global encoder $(E_a^*)$ is saved.


\subsection{Downstream fine-tuning}
During the federated fine-tuning phase, depicted in Figure 2, we initialize the local encoder 
$E_k$
  of the 
 $k_{th}$
  client with the pretrained global encoder 
$E_a$
  acquired from the initial stage. Subsequently, we augment the encoder with a linear downstream task head. The complete model is then fine-tuned using the local labeled data. 
  

\noindent
\textbf{Classification Task} We freeze the weight of pre-trained VIT and add a trainable LoRA adapter, as well as a fully-connected layer, which put on top of pre-trained VIT, for specific classification tasks - Multilabel classification task, Multiclass classification task and binary classification task.

\noindent
\textbf{Segmentation Task} The same pre-trained VIT is applied here for segmentation task. We use of unetr framework\cite{b43} for our evaluation since unet-family allow  pre-trained encoder injected.


\section{Experiments Implementaion}
\subsection{Datasets}










In this work, we use six public fundus dataset in which the first four datasets are used for in-distribution pretraining and task evaluation. The last twos, LACDHS and JSIEC, act as out-of-distribution (ID) test dataset. These two play a role as out-of-network (OOD) new clients joining federated system. We want to evaluate how well own approach generalize to unseen task and OOD data.








\begin{table}[h]
  \begin{center}
    \caption{Statistics of six fundus-image datasets.}
    \label{tab:Statistics of six fundus-image datasets}
    \begin{tabular}{l|c|c|c} 
     \hline
      \textbf{Dataset} & \textbf{Train} & \textbf{Test} & \textbf{Task}\\
      \hline
      RFMiD & 1920 & 640 & Multi-label Classification\\
      DR & 29600 & 6800 & Severity Rating\\
      EYEPACS  & 8000 & 770 & Bianry Classification\\
      REFUGE2 &400  &400 & Optic Disc Segmentation \\
      LACDHS &80  &20 & Vessel Segmentation \\
      JSIEC &800  &200 & Multi-class classification \\
       \hline
    \end{tabular}
    \label{configex}
  \end{center}
\end{table}

\begin{table}[tt]
\centering
\setlength{\tabcolsep}{4pt}
\caption{Results on classification tasks} \label{Classification task results}

\begin{tabular}{lccc}
\hline
{Methods}& Accuracy $\uparrow$ & Precision $\uparrow$  & Macro F1 $\uparrow$ \\
\hline
\multicolumn{3}{l}{$\mathrlap{\textit{Multilabel classification task:}}$} \Tstrut \\
Local Supervision (Split 1)  & 57.5 $\pm$ 0.6  & 13.9 $\pm$ 0.4 & 15.7 $\pm$ 0.3 \\
Local Supervision (Split 2)  & 59.7 $\pm$ 0.3  & 15.3 $\pm$ 0.2 & 15.9 $\pm$ 0.2\\
Centralized SSL (No FL)  & 63.3 $\pm$ 0.5  & 16.9 $\pm$ 0.3 & 17.3 $\pm$ 0.3  \\
SSL-FL (Split 1)  & 58.9 $\pm$ 0.6  & 14.3 $\pm$ 0.5 & 14.8 $\pm$ 0.5 \\
SSL-FL (Split 2)  & 60.9 $\pm$ 0.6  & 15.3 $\pm$ 0.4 & 16.3 $\pm$ 0.3 \\

\hline
$\mathrlap{\textit{Binary classification task:}}$ \Tstrut \\
Local Supervision (Split 1)  & 58.9 $\pm$ 0.4  & 55.3 $\pm$0.2 & 52.1 $\pm$ 0.2 \\
Local Supervision (Split 2)  & 60.4 $\pm$ 0.3  & 54.9 $\pm$ 0.3 & 55.4 $\pm$ 0.4 \\

Centralized SSL (No FL)  & 62.7 $\pm$ 0.3  & 58.7 $\pm$ 0.3 & 56.9 $\pm$ 0.2  \\
SSL-FL (Split 1)  & 59.8 $\pm$ 0.6  & 56.4 $\pm$ 0.3 & 54.2 $\pm$ 0.3 \\

SSL-FL (Split 2)  & 61.1 $\pm$ 0.3  & 57.1 $\pm$ 0.2 & 53.2 $\pm$ 0.2 \\

\hline
$\mathrlap{\textit{Severity rating task:}}$ \Tstrut \\
Local Supervision (Split 1)  & 71.6 $\pm$ 0.3  & 17.2 $\pm$ 0.2 & 17.0 $\pm$ 0.3\\
Local Supervision (Split 2)  & 73.3 $\pm$ 0.1  & 18.7 $\pm$ 0.0 & 18.5 $\pm$ 0.0 \\
Centralized SSL (No FL)  & 73.3 $\pm$ 0.1  & 18.3 $\pm$ 0.0 & 17.9 $\pm$ 0.0  \\
SSL-FL (Split 1)  & 71.5 $\pm$ 0.3  & 16.9 $\pm$ 0.1 & 17.2 $\pm$ 0.2 \\
SSL-FL (Split 2)  & 73.0 $\pm$ 0.3  & 18.4 $\pm$ 0.1 & 18.4 $\pm$ 0.1 \\

\hline

\end{tabular}
\end{table}

\begin{table}[t]
\centering
\caption{Results on segmentation tasks (the lower the better)} \label{tab:metricvsnormalvecs}
\begin{tabular}{lcc}
   Methods & Dice Loss $\downarrow$ & Dice Focal $\downarrow$ \\
\hline
$\mathrlap{\textit{Optic disc segmentation task on REFUGE dataset:}}$ \Tstrut \\
  Local Supervision (Scratch) & 94.9 $\pm$ 0.8 & 76.0 $\pm$ 0.5 \\
  Centralized SSL (No FL) & 92.4 $\pm$ 0.7 & 73.9 $\pm$ 1.2 \\
  SSL-FL (Split 1) & 93.7 $\pm$ 1.1 & \textbf{69.4 $\pm$ 0.8} \\
  SSL-FL (Split 2) & \textbf{91.1 $\pm$ 1.1} & 74.5 $\pm$ 2.3 \\
         \hline

$\mathrlap{\textit{Vessel segmentation task on JSIEC dataset:}}$ \Tstrut \\
  Local Supervision (Scratch) & \textbf{43.9 $\pm$ 0.9} & 64.1 $\pm$ 3.2 \\
  Centralized SSL (No FL) & 44.3 $\pm$ 1.8 & 66.4 $\pm$ 7.2 \\
  SSL-FL (Split 1) & 45.2 $\pm$ 1.7 & 61.4 $\pm$ 1.4 \\
  SSL-FL (Split 2) & 44.7 $\pm$ 1.1 & \textbf{60.7 $\pm$ 1.6} \\
         \hline

\end{tabular}
\end{table}


\begin{figure}[t]
\centering
\subfloat[Split 1]{\label{fig:maeea}\includegraphics[width=.42\linewidth]{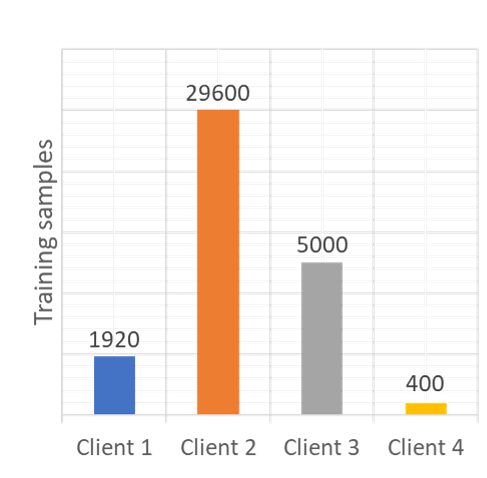}}\hfill
\subfloat[Split 2]{\label{fig:maeeb}\includegraphics[width=.56\linewidth]{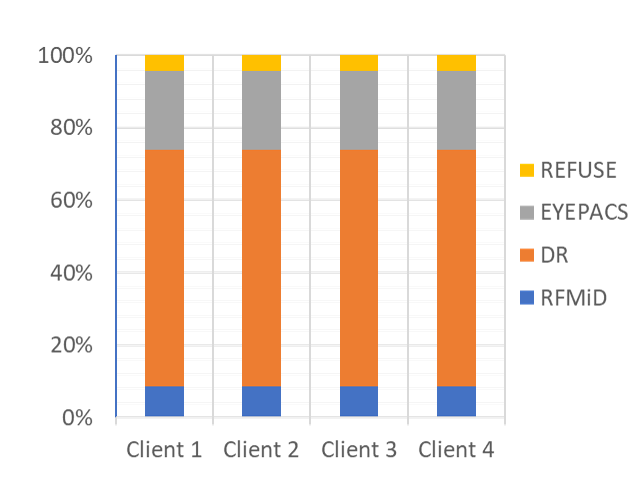}}\hfill
\caption{Data distribution of two split settings. In (a) clients have different tasks and different datasets (modality) while clients in (b) share the same amount of data from each datasets and do the same tasks for such portion of dataset. }
\label{fig:maeevsmae}
\end{figure}

\begin{itemize}

\item  \textit{JSIEC Fundus dataset} 1000 fundus images which belong to 39 classes are come from the Joint Shantou International Eye Centre (JSIEC), Shantou city, Guangdong province, China. 
The copyright of these images belongs to JSIEC.


\item  \textit{RFMiD Small Dataset} consists of 2560 fundus images captured using three different fundus cameras with 46 conditions annotated through adjudicated consensus of two senior retinal experts. 

\item  \textit{EyePACS} The complete Rotterdam EyePACS AIROGS dataset, encompassing both training and testing sets, comprises 8000 color fundus images sourced from subjects across roughly 500 diverse sites with varying ethnic backgrounds.


\item  \textit{REFUSE2} The Retinal Fundus Glaucoma Challenge 2nd Edition (REFUGE2) dataset comprises 800 color fundus images accompanied by annotations for glaucoma classification, optic disc/cup segmentation, and fovea localization.

\item  \textit{DR Dataset} Each image within the Diabetic retinopathy dataset
is assessed by a clinician who assigned a rating on a scale of 0 to 4, corresponding to the absence of diabetic retinopathy (No DR), mild, moderate, severe, and proliferative stages.

\item  \textit{LACDHS Dataset} The dataset comprises 100 fundus digital images of the retina sourced from the Armed Forces Institute of Ophthalmology (AFIO) in Rawalpindi, Pakistan. 
Included within the dataset are annotations detailing the retinal blood vessel network, segmented artery/vein networks utilized for calculating the Arteriovenous Ratio (AVR), as well as annotations of the Optic Nerve Head (ONH). 


\end{itemize}
\subsection{Experimental Settings}
\subsubsection{Baseline} Since we derive a new problem setting, there are no other methods yield to solve it. We used supervised learning on local labels as lower bound for comparison. We assume that in best scenario, data will be visible from all site so that we train a self-supervised encoder from all data and fine-tune it on each tasks to get upper bound performance.
\subsubsection{Task and Data Heterogeneity Setup} We model Task imbalance (Split 1, data split by the dataset) and Task balance (Split 2, data split by drawing same number of data from each dataset ) data distributions of the six datasets, as is shown in \ref{fig:maeevsmae}. Centralized task is the model pretrained on the six dataset together without aggregation. Local supervision task represents pretraining and finetuning on its own task dataset but not federatedly learn the data from other clients.
Simulated data partitions and task partitions allow for a more flexible and thorough investigation of the model behavior, as they can be easily manipulated to test data and task heterogeneity. All the different tasks are finetuned on their own downstream task and the corresponding dataset.

\subsubsection{Self-supervised FL Pre-training and Supervised FL Fine-tuning and Evaluation Metrics} Following \cite{b35}, 
ViT-B \cite{b36} is chosen as the backbone for the proposed models.
Following the setup in MAE\cite{b34}, the input
is split into 16 × 16 patches for MAE. In our
main experiment, we randomly mask at most 60\% of total
image patches for MAE. 
 The downstream model is initialized using the pretrained encoder and fine-tuned with a base learning rate starting at $3e^{-3}$ for all tasks. We use accuracy, precision and F1 score as the evaluation metric for classification on all the datasets.
\section{results and discussion}


\subsection{Results for Classification}

Table \ref{Classification task results} shows the results from different training methodologies across three classification tasks: multilabel classification, binary classification, and severity rating.
In all these classification tasks, the Centralized SSL (No FL) method consistently delivers superior performance, suggesting the advantage for federated learning with lack of data and data communication. The federated learning part shows interesting results. In the multilabel classification task, while SSL-FL (Split 1) shows a decrease in performance compared to centralized SSL, SSL-FL (Split 2) regains some ground, suggesting that the balanced nature of Split 2 may help mitigate some challenges associated with federated settings. In the binary classification task, it is worth to note that both federated learning approaches perform better than local supervision alone, but not much for Split 2, which might due to the great amount of training data for local supervisions in this case, leading to strong representation of the encoder and bringing in information of other dataset does not help much. Still Split 1 with unbalanced data scenarios poses specific challenges that affect performance negatively compared to centralized approaches and Split 2. The severity rating task results reveal a similar pattern where Centralized SSL (No FL) achieves the best overall results. We can see again Split 2 shows improved performance over Split 1, which echoes the trend observed in the multilabel task. Also federated learning does not help much when data is enough for local training. Overall, Centralized SSL provides the best outcomes across all tasks and metrics. The performance differences between SSL-FL splits emphasize the importance of data balancing in federated settings, reducing disparities in model learning and generalization and shows that clients with less data may benefit more from the federated training. 

 \subsection{Results for Segmentation and OOD Tasks}
 \begin{figure}[t]
  \centering
    \includegraphics[width=0.9\linewidth]{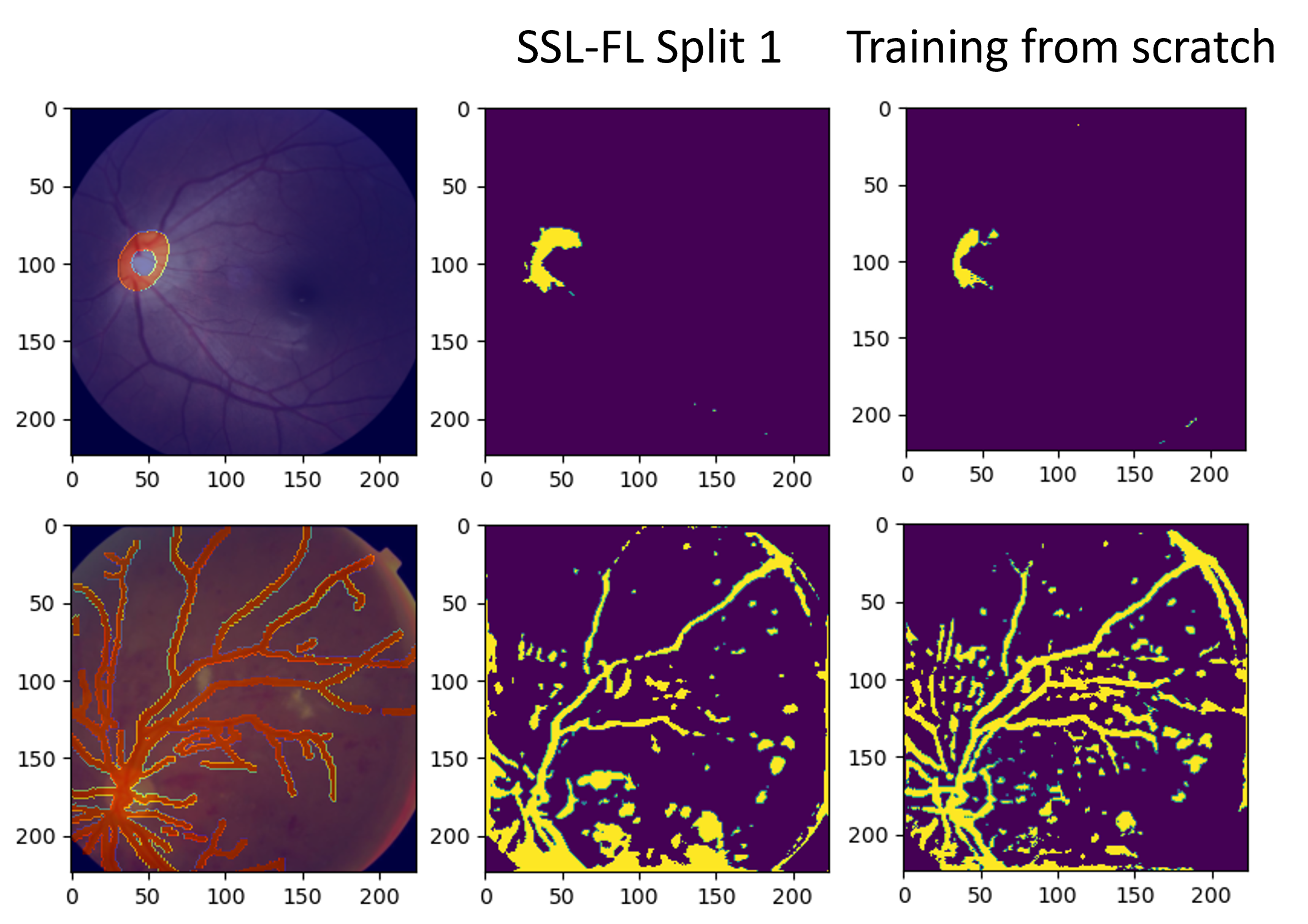}
    \caption{Example of segmentation results given by SSL-FL framework and local supervision.}
    \label{fig:short-b}
\end{figure}

Table \ref{tab:metricvsnormalvecs} presents the results of various methods applied to two different tasks in medical image analysis: optic disc segmentation on the REFUGE dataset and vessel segmentation on the JSIEC dataset. 
For the optic disc segmentation task on the REFUGE dataset, 
the results show that Local Supervision achieves the lowest performance with 94.9\% Dice coefficient for optic disc segmentation and 76.0\% for cup segmentation. Centralized SSL without FL follows closely behind with slightly higher performance. However, when SSL is combined with Federated Learning (SSL-FL), there's a increase in performance, especially noticeable in the cup segmentation metric, with Split 1 achieving the highest scores.
Moving to the vessel segmentation task on the JSIEC dataset, similar trends are observed. Local Supervision achieves the highest Dice coefficients for vessel and background segmentation, with Centralized SSL without FL trailing slightly behind. However, when SSL is coupled with Federated Learning (SSL-FL), there's again a increse in performance, particularly evident in vessel segmentation for Split 2 because of task balance between clients. These results suggest that while SSL can enhance segmentation performance, the introduction of federated learning in this context might not consistently improve results.
The choice of data split also influences the effectiveness of SSL with Federated Learning.


\section*{Conclusion}
In this paper, we propose a privacy-preserving and federated
self-supervised learning framework that collaboratively trains
models on decentralized data using masked image modeling as the heterogeneous self-supervised tasks, in which Low Rank Adaptation(LoRA) is used in the fine tuning part. Our framework is robust to non-IID data distribution across clients, and performs well under severe task heterogeneity and data imbalance settings across diverse medical datasets. Experiments show that while tasks(clients) with less data benefits more from federated pretraining, all with different downstream task will perform better than its local supervision.

\vspace{12pt}

\end{document}